\documentclass[letterpaper]{easychair}

\usepackage{doc}

\usepackage{epstopdf}

\title{Accelerated graph-based nonlinear denoising filters}

\titlerunning{Accelerated graph-based denoising}

\author{
    Andrew Knyazev\inst{1}
\and
    Alexander Malyshev\inst{2}\thanks{The work was supported by Mitsubishi Electric Research Laboratories}\\
}

\institute{
  Mitsubishi Electric Research Laboratories,
  201 Broadway, Cambridge, MA 02139, USA\\
  \email{knyazev@merl.com}
\and
   University of Bergen,
   Department of Mathematics, PB 7803, 5020 Bergen, Norway\\
   \email{alexander.malyshev@math.uib.no}\\
}

\authorrunning{Knyazev and Malyshev}

\begin{document}

\maketitle


\begin{abstract}
Denoising filters, such as bilateral, guided, and total variation filters,
applied to images on general graphs may require repeated application
if noise is not small enough. We formulate two acceleration techniques of
the resulted iterations: conjugate gradient method and Nesterov's acceleration.
We numerically show efficiency of the accelerated nonlinear filters for image denoising
and demonstrate 2-12 times speed-up, i.e., the acceleration techniques reduce the number of
iterations required to reach a given peak signal-to-noise ratio (PSNR) by the above indicated
factor of 2-12.
\end{abstract}


%
%


\section{Introduction}
\label{sec:intro}

Modern image denoising algorithms are edge preserving, i.e., they preserve important
discontinuities in an image while attenuating noise. Many of such algorithms are
based on the anisotropic diffusion idea, first formulated in \cite{PM87,PM90}. The idea
consists in using diffusion coefficients depending on the local variance---the larger
the variance the smaller the coefficient.

Popular denoising techniques, which implement the anisotropic diffusion, include the bilateral filter \cite{AW95,SB97,TM98,E02,PKTD09}, the guided image filter \cite{HST13}, and the total variation denoising \cite{ROF92}. The fastest computer implementations of the bilateral filter are proposed in recent papers
\cite{C13,SK15}. The guided image filter has been included in the MATLAB Image Processing Toolbox.
We also remark that the total variation denoising can be formulated in the filter form; see Section~\ref{ssec:tv}. All the three filters may be applied to images or signals on graphs; see, e.g., \cite{LG12} on the graph-based methods in signal and image processing.

More recent state-of-the-art denoising methods are patch-based such as those developed
in \cite{BCM05,KB06,DFKE07,CM12,LCBM12}. An improvement of these methods, based on a special truncation
of high frequency modes, is proposed in \cite{TM14a,TM14b}. Since the patch-based algorithms
use geometrically similar patches, they seem to be inconvenient for images or signals on general graphs.
This reason might partially justify a still active research interest in the basic imaging techniques like the bilateral filter, guided image filter and total variation denoising. Moreover, the models based on the total variation enjoy very rich variational properties.

In certain situations, a single application of a smoothing filter does not produce an acceptable denoising result, and, therefore, the filter transform has to be applied repeatedly (or iteratively), say 10-1000 times, depending on the filter type and level of noise. The repetitive application procedure may be expensive even for images of moderate size. Our previous work in \cite{KM15a,KM15b} is devoted to acceleration techniques for the iterative application of smoothing filters formulated above. The results are based on the studies in \cite{GNO13,TKMV14}, where low-pass filters are constructed by means of projection onto the leading invariant subspaces, corresponding to the modes of lowest frequency, of a graph Laplacian matrix generated by a basic smoothing filter.

The initial publications \cite{GNO13,TKMV14,KM15a} consider iterative application of a fixed
smoothing filter, whose coefficients are defined by the input noisy image. Such a method is
known under the name of power iteration. The authors of \cite{GNO13} propose to accelerate
the power iteration by the aid of Chebyshev's polynomials. The paper \cite{TKMV14} additionally proposes to accelerate the power iteration by the aid of the polynomials generated in the conjugate gradient method \cite{V03}. In \cite{KM15a}, we formulate a special variant of the preconditioned conjugate method,
which accelerates the power iteration for 1D and 2D signals on graphs, and demonstrate that similar
acceleration can be achieved with the LOBPCG method \cite{K01}.

The subsequent works \cite{KM15b,SOIT15} deal with a nonlinear iterative application of filters, where a smoothing filter at each iteration is determined by the currently processed image. The resulting transform yields a nonlinear smoothing filter in contrast to the linear smoothing filter given by the power iteration with a fixed filter at each iteration. The paper \cite{KM15b} presents a special variant of a nonlinear preconditioned conjugate gradient method and numerically demonstrates its high efficiency for accelerated denoising of one-dimensional signals. The conference presentation \cite{SOIT15} shows how to accelerate the nonlinear iterative filters by means of the Chebyshev polynomials.

The present note continues the work in our previous papers \cite{KM15a,KM15b} about acceleration of iterative smoothing filters and contains a number of new contributions listed below. In addition to the bilateral and guided image filters, we consider the total variation denoising and formulate it as a filter operator. In addition to the preconditioned conjugate gradient (PCG) acceleration of nonlinear iterative smoothing filters, we propose to apply Nesterov's acceleration, which is commonly used in a totally different context of iterative solution of convex minimization problems. We numerically investigate performance of the PCG acceleration of nonlinear iterative smoothing filters for two-dimensional images, which is not clear from the previous publications at all. We also numerically investigate performance of Nesterov's acceleration of nonlinear iterative smoothing filters.

\section{Smoothing filters}
\label{sec:filt}

We consider only smoothing filters, which are represented in the matrix form $x^1=D^{-1}Wx^0$,
where the vectors $x^0$ and $x^1$ of length $N$ are the input and output signals,
respectively. The entries $w_{ij}$ of the symmetric $N\times N$ matrix $W$ are determined
by a guidance signal $g$, i.e. $W=W(g)$. When $g=x^0$, the filter is nonlinear and called self-guided.
The diagonal matrix $D$ has $N$ positive diagonal entries $d_i=\sum_{j=1}^Nw_{ij}$. The symmetric nonnegative definite matrix $L=D-W$ is commonly referred to as a graph Laplacian matrix. The spectrum of the normalized Laplacian $D^{-1/2}LD^{-1/2}$ is nonnegative real, and its largest eigenvalues correspond
to the highest oscillation modes.

In this paper, we are interested in iterative application of the filter transform
\[
x^{k+1}=D^{-1}Wx^k,\qquad k=1,\dots,k_{\max},
\]
where the number of iterations $k_{\max}$ needs to be chosen large enough for good denoising result in $x^{k_{\max}}$, but small enough to prevent over-smoothing effect. Each iteration $k$ is a self-guiding filter, where the weights $w_{ij}$ are determined by the guidance signal $g=x^k$.

\subsection{Bilateral filter (BF)}
\label{ssec:bf}

Let us assume that a spatial distance $\|p_i-p_j\|\in[0,\infty]$ can be defined for all index pairs $(i,j)$, $1\leq i,j\leq N$. The weights of the bilateral filter \cite{TM98} equal
\[
 w_{ij}=\exp\left(-\frac{\|p_i-p_j\|^2}{2\sigma_d^2}\right)
 \exp\left(-\frac{|g_i-g_j|^2}{2\sigma_r^2}\right),
\]
where the constants $\sigma_d$ and $\sigma_r$ are filter parameters, and $|g_i-g_j|$ is
a suitable distance between the components of a guidance signal $g$.
The arithmetical complexity of a single application of the bilateral filter to
images on rectangular grids can be reduced to $O(N)$;
see \cite{C13}.

\subsection{Guided filter (GF)}
\label{ssec:gf}

The following algorithm, proposed in \cite{HST13} and implemented in the MATLAB Image Processing Toolbox, performs one application of the guided filter defined by a guidance signal $g$:

\begin{tabular}{l}
\hline\\[-2ex]
\textbf{Algorithm}{ Guided filter}\\\hline\\[-2ex]
\textbf{Input:} $x$, $g$, $w$, $\epsilon$\\
\textbf{Output:} $y$\\
\quad $mean_g=f_{mean}(g,w)$;\quad $mean_x=f_{mean}(x,w)$\\
\quad $corr_g=f_{mean}(g.*g,w)$;\quad $corr_{gx}=f_{mean}(g.*x,w)$\\
\quad $var_g=corr_g-mean_g.*mean_g$;\quad $cov_{gx}=corr_{gx}-mean_{g}.*mean_x$\\
\quad $a = cov_{gx}./(var_g+\epsilon)$;\quad $b = mean_x-a.*mean_g$\\
\quad $mean_a=f_{mean}(a,w)$;\quad $mean_b=f_{mean}(b,w)$\\
\quad $y=mean_a.*g+mean_b$\\\hline\\[-2ex]
\end{tabular}

\noindent The function $f_{mean}(\cdot,w)$ denotes a mean filter with the window width $w$.
The positive constant $\epsilon$ determines the smoothness degree---the larger $\epsilon$
the larger smoothing effect. The dot preceded operations $.*$ and $./$ denote the componentwise
multiplication and division of vectors or matrices. Special implementations of the guided filter
applied to images on rectangular grids achieve an $O(N)$ arithmetical complexity;
see \cite{HST13}.

The above algorithm does not explicitly build the matrices $W$ and $D$ for the filter transform
$y=D^{-1}Wx$. Nevertheless, the transform is linear, and its matrix coincides with $W$,
when boundary padding for the mean filter $f_{mean}(\cdot,w)$ is defined carefully
so that the matrix $W(g)$ would be symmetric. Thus, the diagonal matrix $D$ for the guided filter
equals the identity matrix $I$ \cite{HST13}, and the graph Laplacian matrix is $L=I-W$.

\subsection{Total Variation filter (TVF)}
\label{ssec:tv}

Let $\nabla$ denote a gradient operator acting on signals.
The classical Rudin-Osher-Fatemi (ROF) denoising model \cite{ROF92} reads as follows:
\[
\min_x\|\nabla x\|_1 \mbox{ subject to } \|x-x^0\|_2=\sigma,
\]
where $\sigma>0$ is given. Solution of ROF for a two-dimensional continuous image $x^0(x,y)$ is approximated by the solution $u(x,y,t)$ of the following boundary value problem
for sufficiently large $t>0$, a suitable constant $\lambda>0$, and sufficiently small
regularizing parameter $\epsilon>0$:
\begin{align*}
&u_t=\frac{\partial}{\partial x}\left(\frac{u_x}{\sqrt{u_x^2+u_y^2}+\epsilon}\right)+
\frac{\partial}{\partial y}\left(\frac{u_y}{\sqrt{u_x^2+u_y^2}+\epsilon}\right)-\lambda(u-x^0)\\
&u(x,y,0)=x^0(x,y);\quad
\frac{\partial u}{\partial n}=0\text{ on the image boundary}.
\end{align*}
The value $\|\nabla u\|_1=\iint\sqrt{u_x^2+u_y^2}dxdy$ is called the total variation of $u(x,y)$.

The graph Laplacian matrix associated with the ROF model is a suitable discretization
of the elliptic operator
$-\frac{\partial}{\partial x}\left(\frac{u_x}{\sqrt{u_x^2+u_y^2}+\epsilon}\right)-
\frac{\partial}{\partial y}\left(\frac{u_y}{\sqrt{u_x^2+u_y^2}+\epsilon}\right)$
with the Neumann boundary conditions.

For a one-dimensional $N\times1$ array $x$, we can, e.g., define the gradient operator $\nabla x=Gx$ by means of the bidiagonal $N\times N$ matrix
\[
G=\begin{bmatrix}-1&1\\&\ddots&\ddots\\&&-1&1\\&&&0\end{bmatrix}.
\]
Given a one-dimensional guidance signal $g$ and regularization parameter $\epsilon>0$, we introduce a
diagonal diffusion $N\times N$ matrix $\mbox{diag}(C)$ with the diagonal
$C=\frac{1}{4}\left[\epsilon./(\epsilon+|\nabla g|)\right]$.
The TV filter $y=D^{-1}Wx$ is then defined by the $N\times N$ matrices
$L(g)=\nabla^T\mbox{diag}(C)\nabla=G^T\mbox{diag}(C)G$, $D(g)=I$, $W(g)=D-L$.

The gradient operator applied to a two-dimensional $M\times M$ array $x$ can be defined by the formula
$\nabla x=\begin{bmatrix}p_1\\p_2\end{bmatrix}=\begin{bmatrix}Gx\\xG^T\end{bmatrix}$ with the bidiagonal $M\times M$ matrix $G$. The transposed gradient is the operator
$\nabla^T\begin{bmatrix}p_1\\p_2\end{bmatrix}=G^Tp_1+p_2G$.
For an $M\times M$ guidance array $g$, we use the coefficient matrix
\[
C=\frac{1}{8}\left[\epsilon./\left(\epsilon+
\sqrt{|Gg|.*|Gg|+|gG^T|.*|gG^T|}\right)\right],
\]
which is rotation invariant.
Then the Laplacian operator is $L(g)x=\nabla^T\left(\begin{bmatrix}C\\C\end{bmatrix}.*\nabla x\right)=
G^T(C.*Gx)+(xG^T.*C)G$, $D(g)=I$, and $W(g)=D-L$.

In the case of general graph-based signal $x$ and guidance $g$, one can use the above formulas for the one-dimensional case with a problem specific gradient operator.

\section{Acceleration of iterations}
\label{sec:accel}

In this section, we provide two algorithms for acceleration of the non-linear filtering iteration $x_0=x$, $x_{k+1}=D^{-1}(x_k)W(x_k)x_k=x_k-D^{-1}(x_k)L(x_k)x_k$, $k=0,1,\ldots,k_{\max}$, where the symmetric matrices $W(g)$ and $D(g)$ are defined in Section~\ref{sec:filt}, and $L(g)=D(g)-W(g)$.
In both algorithms, the total running time is determined by the number of calls to the
basic filter $x_{k+1}=D^{-1}(x_k)W(x_k)x_k$. The overhead due to the auxiliary computations
for the accelerations is marginal with respect to the application of the basic filters
to images on general graphs.

\subsection{Acceleration by the Preconditioned Conjugate Gradient (PCG) method}
\label{ssec:pcg}

Formally applying the classical preconditioned conjugate gradient method \cite{V03}
with a preconditioner $D$ to the system of linear equations $Lu=0$,
and choosing $D$ and $W$ depending on the current approximation $y$ to the solution,
we arrive at the following algorithm, proposed and partially tested in \cite{TKMV14,KM15a,KM15b}.
We draw attention to the necessity of restarts in the algorithm due to
the nonlinearity of power iterations. We also emphasize that convergence theory of PCG
in the considered context does not exist.

\vspace{1ex}
\begin{tabular}{l}
\hline\\[-2ex]
\textbf{Algorithm}{ PCG($k_{\max}$) with $l_{\max}$ restarts}\\\hline\\[-2ex]
\textbf{Input:} $x$, $k_{\max}$, $l_{max}$\\
\textbf{Output:} $y$\\
$y=x$\\
\textbf{for }{$l=1,\ldots,l_{\max}$}\textbf{ do}\\
\quad $r=W(y)y-D(y)y$\\
\quad\textbf{for }{$k=1,\ldots,k_{\max}-1$}\textbf{ do}\\
\qquad $s=D^{-1}(y)r$\\
\qquad $\gamma=s^Tr$\\
\qquad\textbf{if }$k=1$\textbf{ then} $p=s$\\
\qquad\textbf{else} $\beta=\gamma/\gamma_{old}$; $p=s+\beta p$\\
\qquad\textbf{endif}\\
\qquad $q=D(y)p-W(y)p$\\
\qquad $\alpha=\gamma/(p^Tq)$\\
\qquad $y=y+\alpha p$\\
\qquad $r=r-\alpha q$\\
\qquad $\gamma_{old}=\gamma$\\
\quad\textbf{endfor}\\
\textbf{endfor}\\\hline\\[-2ex]
\end{tabular}

\subsection{Nesterov's acceleration}
\label{ssec:na}

Nesterov's acceleration is suggested in \cite{N83}. The choice of $\beta$ in the
following algorithm has been adopted from \cite{SBC14}. To our best knowledge,
convergence theory of Nesterov's acceleration in the present context is not available.

\vspace{1ex}
\begin{tabular}{l}
\hline\\[-2ex]
\textbf{Algorithm}{ Nesterov($k_{\max}$)}\\\hline\\[-2ex]
\textbf{Input:} $x$, $k_{\max}$\\
\textbf{Output:} $y$\\
$y=x$; $y_{old}=y$\\
\textbf{for }{$k=1,\ldots,k_{\max}$}\textbf{ do}\\
\quad $\beta=(k-1)/(k+2)$\\
\quad $t=y+\beta(y-y_{old})$\\
\quad $y_{old}=y$\\
\quad $y=D^{-1}(t)W(t)t$\\
\textbf{endfor}\\\hline\\[-2ex]
\end{tabular}

\section{Numerical study}
\label{sec:numer}

From the mathematical point of view, application of the smoothing filter transform $x^1=D^{-1}Wx^0$
to images on general graphs does not differ from the case of images on rectangular grids,
if specific geometric properties of the rectangular grid on the plane are not taken into account.
Therefore, in order to facilitate programming efforts in our numerical tests, we carry out numerical experiments with images on standard rectangular grids instead of images on general graphs.
We use a gray-scale $512\times512$-image created by the MATLAB command
\texttt{clean = phantom('Modified Shepp-Logan',512)}. The image is piecewise constant,
and its intensity levels span the range $[0,1]$.
A noisy image, generated by the MATLAB command \texttt{noisy = imnoise(clean)},
is corrupted by Gaussian white noise with zero mean and variance equal to $0.01$.
The peak signal-to-noise ratio (PSNR) of the noisy image is $21.7$. In order to show
smaller details, we display the zoomed image patches consisting of the rows 211:420 and columns 201:300
instead of full $512\times512$-images. But the filters are applied to full images, and
the reported PSNR values are also computed for full images.

Our simple implementation of the bilateral filter has the the following parameters:
the window width equals 5, $\sigma_d=1$, and $\sigma_r=0.2$.
As the guided image filter, we use the function \texttt{imguidedfilter} from MATLAB
with the window width 5 and the smoothness parameter $\epsilon=0.0001$.
The regularization parameter in our implementation of the total variation filter
equals $\epsilon=0.001$. The restart parameter in the preconditioned conjugate gradient
acceleration is $k_{\max}=3$. According to our experience, the PCG acceleration of
self-guided smoothing filters with $k_{\max}>5$ does not work well.
We count the total number of iterations in the iterative filters as
the number of calls of a basic filter. Therefore, the number of iterations in the PCG method
equals $k_{\max}\times l_{\max}$.

\begin{figure}
\centering
\begin{minipage}{.45\linewidth}
\includegraphics[width=\textwidth]{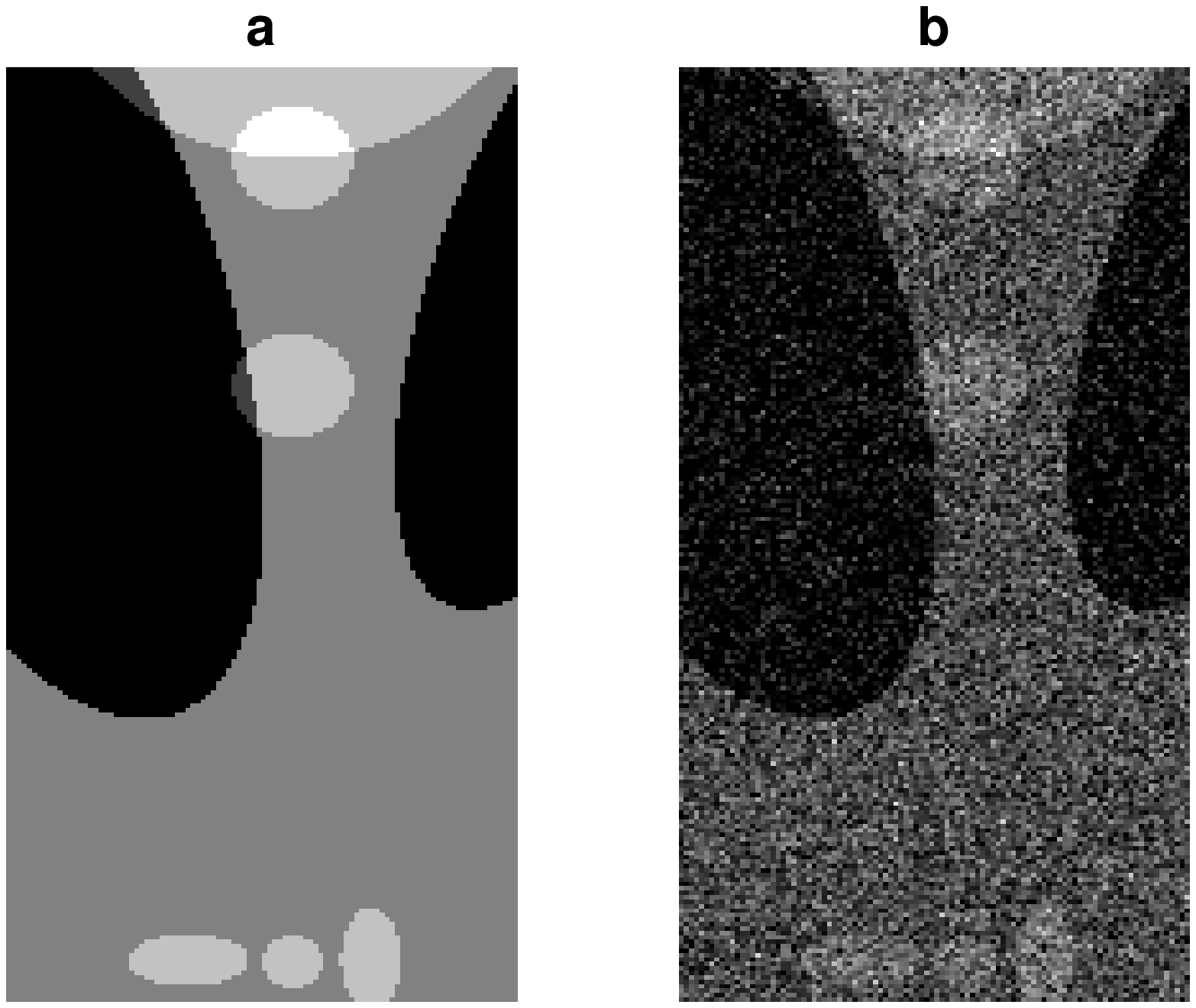}
\caption{Clean image vs. noisy image, PSNR = $21.72$}
\vspace*{5.6ex}
\label{fig1}
\end{minipage}
\hspace{.05\linewidth}
\begin{minipage}{.45\linewidth}
\includegraphics[width=\textwidth]{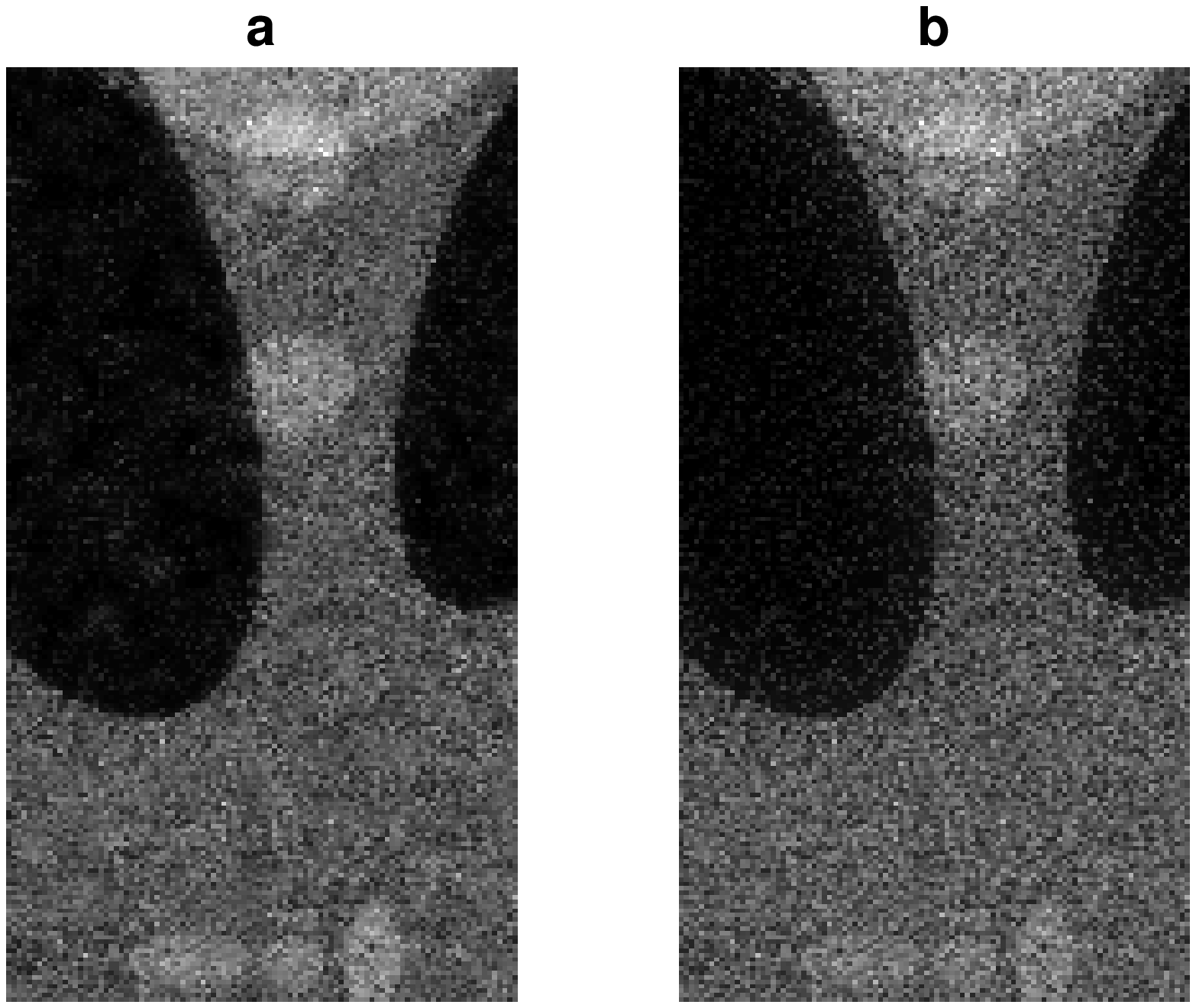}
\caption{A single application of the guided filter with $\epsilon=0.01$.
(a) window width~$5$, PSNR=26.16 vs. (b) window width $30$, PSNR=24.84}
\label{fig2}
\end{minipage}
\end{figure}

Figure~\ref{fig1} displays zooms of the clean and noisy images. Figure~\ref{fig2}
shows the denoising result of a single application of the guided image filter with the
default MATLAB's value of the smoothness parameter $\epsilon=0.01$.
The image on the left has been processed with the window width 5, which is the MATLAB's default value.
The image on the right has been processed with the window width 30.

\begin{figure}
\centering
\begin{minipage}{.75\linewidth}
\includegraphics[width=\textwidth,height=.55\textwidth]{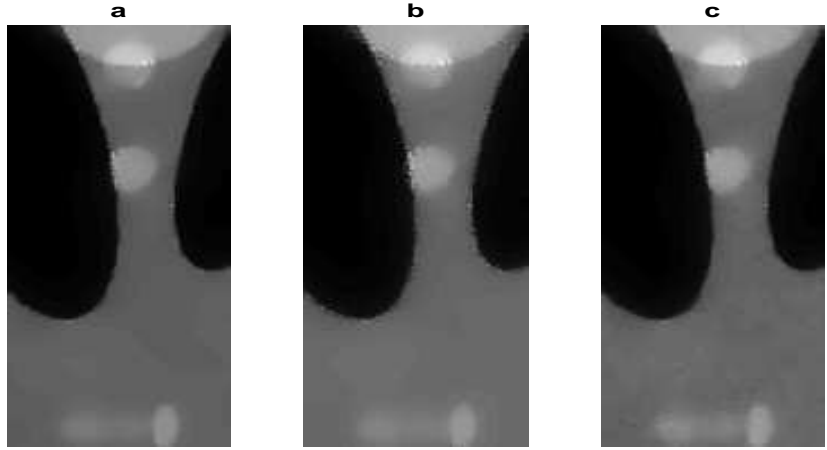}
\caption{(a) 70 iterations of the repeated guided filter, PSNR=29.13, (b) 30 iterations of the PCG
accelerated guided filter, PSNR=28.76,\newline (c) 23 iterations of the Nesterov accelerated guided filter, PSNR=29.01}
\label{fig3}
\end{minipage}
\end{figure}

\begin{figure}
\centering
\begin{minipage}{.75\linewidth}
\includegraphics[width=\textwidth,height=.55\textwidth]{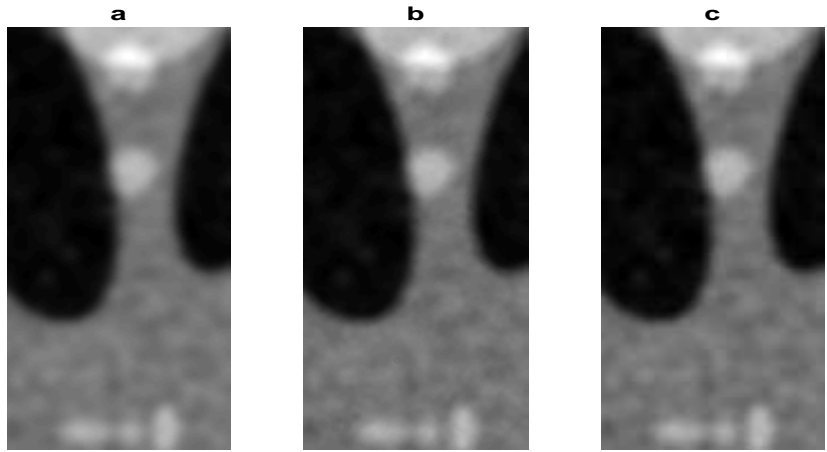}
\caption{(a) 10 iterations of the repeated bilateral filter, PSNR=29.69, (b) 6 iterations of the PCG
accelerated bilateral filter, PSNR=29.82, (c) 5 iterations of the Nesterov accelerated bilateral filter, PSNR=29.85}
\label{fig4}
\end{minipage}
\end{figure}

\begin{figure}
\centering
\begin{minipage}{.75\linewidth}
\includegraphics[width=\textwidth,height=.55\textwidth]{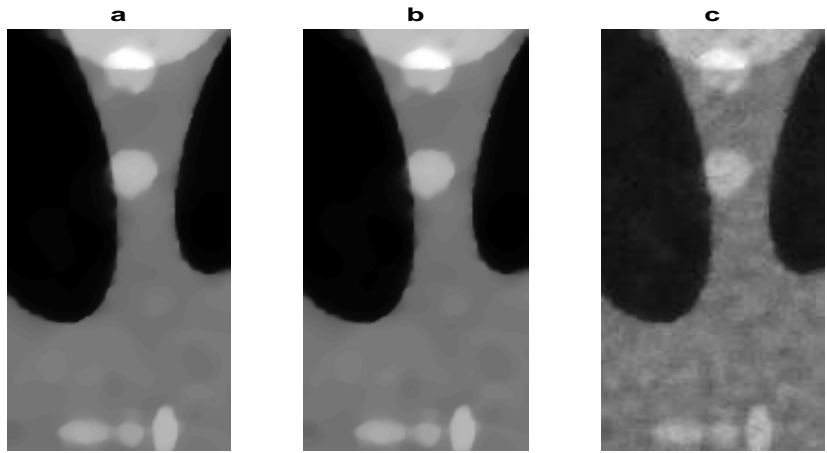}
\caption{(a) 1000 iterations of the repeated TV filter, PSNR=28.50, (b) 135 iterations of the PCG accelerated TV filter, PSNR=28.48,\newline (c) 80 iterations of the Nesterov accelerated TV filter, PSNR=28.31}
\label{fig5}
\end{minipage}
\end{figure}

Figures~\ref{fig3}--\ref{fig5} illustrate the repeated application,
the PCG accelerated iteration, and the Nesterov accelerated iteration,
of the guided filter, the bilateral filter,
and the total variation filter, respectively.
We have chosen the filter parameters and number of iterations in order
to reach sufficiently good visual quality of output images.
We remark that selection of output results with the highest possible PSNR values is not always
the best strategy for achieving the best visual quality.

The results in Figures \ref{fig3} and \ref{fig4} show 2-3x speedup for the accelerated
guided image and bilateral filters. The speedup for the accelerated total variation filter
shown in Figure~\ref{fig5} is 8-12x. The visual quality of the images \ref{fig5}(a) and
\ref{fig5}(b) is slightly better than that of \ref{fig5}(c).
As concerns comparison of the two acceleration techniques by the preconditioned conjugate
gradient method and by Nesterov's acceleration, both methods usually possess similar speedup
and output quality. However, sometimes Nesterov's method
behaves stabler in the acceleration of nonlinear iterations. The PCG acceleration
is more efficient for acceleration of linear iterations.

\begin{figure}
\centering
\begin{minipage}{\linewidth}
\includegraphics[width=\textwidth,height=.35\textwidth]{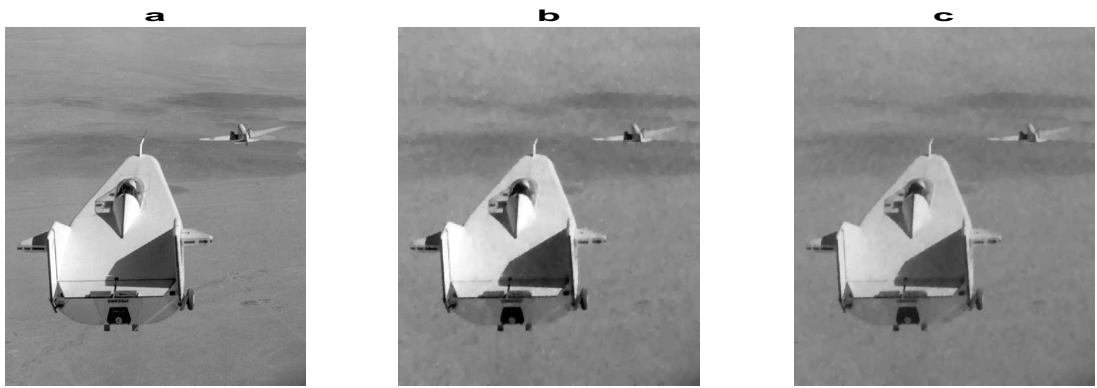}
\caption{(a) clean image, (b) 800 iterations of the repeated TV filter, PSNR=33.18,\newline
(c) 45 iterations of the PCG accelerated TV filter, PSNR=33.02}
\label{fig6}
\end{minipage}
\end{figure}

The preconditioned conjugate gradient acceleration may work especially well for the total variation denoising, when the graph Laplacian matrix is not properly scaled. Figure~\ref{fig6} shows results of
the PCG accelerated TV denoising for the $512\times 512$-image \texttt{liftingbody.png} from MATLAB, corrupted by Gaussian noise with zero mean and variance 0.01. The PSNR value of the noisy image is 20.06.

There exists a variant of acceleration proposed in \cite{P64}, which is often called the heavy ball method. According to our numerical experience, we can conclude that the acceleration based on the heavy ball method is inferior to Nesterov's acceleration.

We have also carried out extensive experiments with the BM3D code, developed by the authors
of the BM3D method \cite{DFKE07} and currently considered as one of the best patch-based
denoising codes. BM3D with the default parameters usually requires only one application to reach
the best possible quality. With other acceptable parameter choices, BM3D produces the best quality
after 1 or 2 iterations. It means that our accelerations techniques are of no use for the
BM3D code. However, we believe that our accelerations may be very useful for denoising signals and images
on general graphs, where the patch-based methods do not work.

\section{Conclusion}
\label{sec:conclusion}

We have numerically demonstrated that acceleration by the preconditioned conjugate gradient
and Nesterov's methods works for the iterative self-guided smoothing filters in the 2D case.
Both acceleration methods demonstrate similar efficiency. The accelerated filters achieve
approximately the same denoising quality,
e.g. in terms of PSNR, as the non-accelerated iterative application of basic filters such as the bilateral, guided, and total variation filters. The acceleration speedup, measured by the number of calls to basic smoothing filters, on images of moderate size, say $512\times512$, can be in the range 2-12x.
We remark that mathematical theory of the proposed accelerations of nonlinear smoothing filters
is not developed yet.
The accelerated filters may be especially useful for processing images and signals on general graphs.


\bibliographystyle{plain}

\end{document}